# S4D-Bio Audio Monitoring of Bone Cement Disintegration in Pulsating Fluid Jet Surgery under Laboratory Conditions

Melanie Schaller (0000-0002-5708-4394), Sergej Hloch (0000-0003-4066-0620),
Akash Nag (0000-0003-0400-7739), Dagmar Klichova (0000-0002-7135-7141),
Nick Janssen (0009-0006-4372-8466), Frank Pude (0009-0005-5837-5037),
Michal Zelenak (0000-0002-3066-3097), Bodo Rosenhahn (0000-0003-3861-1424)

*Abstract*—This study investigates a pulsating fluid jet as a novel precise, minimally invasive and cold technique for bone cement removal. We utilize the pulsating fluid jet device to remove bone cement from samples designed to mimic clinical conditions (see Figure 1). The effectiveness of long nozzles was tested to enable minimally invasive procedures. Audio signal monitoring, complemented by the State Space Model (SSM) S4D-Bio, was employed to optimize the fluid jet parameters dynamically, addressing challenges like visibility obstruction from splashing (see Figure 1). Within our experiments, we generate a comprehensive dataset correlating various process parameters and their equivalent audio signals to material erosion. The use of SSMs yields precise control over the predictive erosion process, achieving 98.93 % accuracy. The study demonstrates on the one hand, that the pulsating fluid jet device, coupled with advanced audio monitoring techniques, is a highly effective tool for precise bone cement removal. On the other hand, this study presents the first application of SSMs in biomedical surgery technology, marking a significant advancement in the application. This research significantly advances biomedical engineering by integrating machine learning combined with pulsating fluid jet as surgical technology, offering a novel, minimally invasive, cold and adaptive approach for bone cement removal in orthopedic applications.

## I. INTRODUCTION

The removal of bone cement [1] during revision surgeries is a critical task traditionally performed using manual methods [2]. These methods can be time-consuming, imprecise, and potentially damaging for surrounding materials [3]. Manual removal often requires considerable skill and poses significant challenges, especially when aiming to preserve the integrity of the surrounding bone for future implant fixation [3, 4]. Studies have shown that traditional methods can be quite destructive, emphasizing the need for a more refined approach [5].

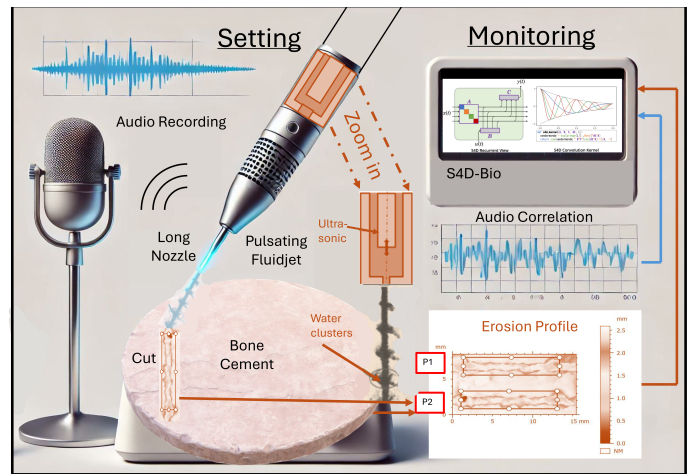

Fig. 1. Setup and monitoring of bone cement removal using a pulsating fluid jet with a long nozzle. The process utilizes audio recording, with a microphone capturing sound signals generated during erosion. The fluid jet, pulsated with ultrasonic waves forms clusters of fluid for higher erosion effects. Monitoring is conducted via the S4D-Bio system, which correlates audio data with erosion profiles to analyze material removal depth and efficiency according to the standoff distance.

In orthopedic surgeries, approximately 19% of patients who receive an implant will require a reoperative revision surgery within 15-23 years [6]. This underlines the pressing need for techniques that minimize damage during such procedures. This signifies the necessity for improved tools and procedures that can enhance surgical outcomes and reduce the risk of further complications.

In this work, we introduce a novel approach using a pulsating fluid jet device as an alternative to conventional techniques [7, 8] (see Figure 1). This method offers substantial advantages, including a controlled and cold method of removing bone cement, thereby potentially reducing surgery

M. Schaller, N. Janssen, and B. Rosenhahn are with the Institute for Information Processing, Leibniz University Hannover, 30167 Hannover (e-mail: schaller@tnt.uni-hannover.de).
S. Hloch, A. Nag, M. Zelenak and D. Klichova are with Czech Academy of Sciences, Institute of Geonics, Ostrava, Czech Republic.
S. Hloch and A. Nag are with Faculty of Mechanical Engineering, VŠB TU Ostrava, Czech Republic.
S. Hloch is with Faculty of Manufacturing Technologies, TUKE, Prešov, Slovakia.
F. Pude is with Study Center for IT-Management & Computer Science (SZI), Baden-Wuerttemberg Cooperative State University, Loerrach, Germany and Steinbeis Consulting Center for High-Pressure Waterjet Technology, Horgau, Germany.

time and tissue damage [9]. However, the challenge remains in tailoring the parameters of the fluid jet to the individual bone cement characteristics. Additionally, real-time monitoring of the fluid jet's behavior is critical, as traditional monitoring methods often fail due to splashing effects that obscure visual feedback. Therefore, we propose the use of audio signals [10, 11]as novel monitoring technique in this field to optimize the fluid jet parameters during bone cement removal [12] (see Figure 1).

Despite promising results from the use of pulsating fluid jet devices [13, 8], there remains a significant challenge in optimizing the fluid jet process parameters specifically for bone cement removal. Current studies do not adequately address the use of long nozzles required for minimally invasive approaches (see Figure 1), nor do they explore the efficiency of audio monitoring in real surgical settings. This lack of research highlights the necessity for precise control over the erosion process while maintaining integrity.

To address these gaps, our research investigates the efficacy of long nozzles and to develop robust online monitoring techniques. The focus lies in estimating the safe erosion zone within the standoff distance using a stairs trajectory method to detect changes in both erosion levels and the corresponding audio signal. The standoff distance is the space between the nozzle of the fluid jet device and the surface of the bone cement, that effects the precision of material removal during the procedure. Our approach defends and rationalizes the use of specific measurement tools and methodologies.

For the experimental part, bone cement samples are prepared with intentionally uneven surfaces to simulate realistic conditions [14]. These samples are securely clamped to ensure reliability and consistency during testing.

Traditionally, cement removal has been carried out using methods described as "destructive," underlining the potential of the fluid jet as a less invasive alternative. The fluid jet should be configured on a per-case basis, with attention to factors like standoff distance. We propose audio signal analysis to optimally adjust the fluid jet's parameters in real-time, employing machine learning techniques to navigate the complex physical process underlying these adjustments. State Space Models [15] are explored as a promising method to correlate audio feedback with fluid jet settings.

Our main contributions are outlined as follows:
- We propose a pulsating fluid jet device to carefully remove bone cement as an alternative to manual procedures.
- The challenge of limited visual feedback during the cutting process due to splashing is addressed by using audio monitoring.
- The pulsating fluid jet device offers several degrees of freedom. We conduct multiple experiments to generate a dataset of different standoff distances, observing their effects on materials with the associated audio signals.
- Audio signals are utilized to monitor the parameter variance. Through machine learning, using State Space Models for time series analysis, we achieve effective process control. This study represents the first application of SSMs in biomedical surgery technology, marking a pioneering advancement in the field.
- We make the dataset publicly available under IEEE Dataport and deliver the scripts to load the data as well as preprocessing scripts[1].

The model predictions are first validated under laboratory conditions using measured profiles as ground truth, demonstrating high accuracy and suggesting potential for real-world application in industries and surgeries.

## II. RELATED WORK

In the following subsections, we divide the related work into the biomedical engineering perspective and the machine learning perspective.

### A. Related work in Biomedical Engineering

Polymethylmethacrylate (PMMA) is a synthetic biomaterial that has been used in orthopedic surgeries since the 1940s for replacing joint surfaces and/or repairing fractures [16]. The disintegration of bone cement, particularly PMMA, is a significant concern in orthopedic surgeries such as joint replacements [17]. This degradation can directly affect surgical outcomes and patient satisfaction [18]. Current methods for the removal of PMMA during revision total joint arthroplasty include the cement-on-cement technique, specialized instruments, thermomechanical methods, and biomechanical approaches. The cement-on-cement method has demonstrated a 91% success rate in hip and knee revision surgeries, notably in septic cases [19]. However, these traditional techniques often require specialized instruments and access enhancements, such as extended osteotomies, to ensure complete cement removal. These enhancements can increase the complexity of the surgical procedure and the risk of complications. Thermomechanical methods, which use heat to modify the properties of the bone cement, facilitate its removal by reducing the force necessary for detachment. Despite their effectiveness, these methods can pose the risk of overheating and damaging surrounding tissues, particularly when utilizing ultrasonic or laser devices [20, 4]. The choice of removal technique is often influenced by patient-specific factors and the particular challenges encountered during surgery.

The introduction of fluid jet technology into orthopedic applications offers several advantages, including the cold and targeted disintegration of bone cement without damaging adjacent tissues [13, 9, 21]. Pulsating fluid jets, in particular, have emerged as a promising, minimally invasive method for bone cement removal [7, 8]. The effectiveness of this technique is influenced by various factors such as nozzle diameter, standoff distance, and thepressure of the fluid, all of which require optimization for successful application [7, 12]. While research suggests that jet-lavage techniques could potentially improve cement penetration and bonding in total knee arthroplasties [22], the adoption of pulsating fluid jets in clinical practice still faces difficulties. These challenges include optimizing parameters like standoff distance, flow rate, frequency, and acoustic chamber length to maximize wear rates for different cement types [23, 24, 25].

---
[1]https://ieee-dataport.org/documents/bone-cement-removal-audio-monitoring-and-erosion-depth

Recent advancements in imaging and sensor technologies have made it possible to monitor bone cement erosion with greater precision during surgery. Bone cement pulsating fluid jet erosion monitoring provides innovative solutions to the limitations associated with conventional cement removal techniques [12, 26, 27]. Understanding and monitoring this erosion process is crucial for enhancing surgical practices and ensuring long-lasting patient outcomes [28]. Thus acoustic monitoring, facilitated by these technological advancements, could provide critical insights during and post-surgery, improving the precision and efficacy of orthopedic procedures.

### B. Related work in Machine Learning

Integrating physical processes into neural network models enriches our understanding of dynamic systems by effectively capturing temporal dependencies in real-world data [29, 30]. This aligns with the rising use of State Space Models in deep learning for dynamic system representation [31, 32, 15, 33, 34], offering an alternative to the widely used Transformer Architectures [35]. The S4D model, renowned for its optimized sequence modeling architecture, is adapted as the foundational framework within the S4D-Bio model, effectively extending its capabilities to real-time data analysis in the context of biomedical engineering.

For benchmarking, we utilize Long Short-Term Memory (LSTM) networks [36], valued for their capacity to manage long-term dependencies, and Gated Recurrent Units (GRUs) [37], which provide efficient training with fewer parameters. Additionally, we compare performance with Multi-Layer Perceptrons (MLP) [38], which is a well-established architectural approach for capturing non-linear data patterns. To enhance the adaptability and expressiveness of the MLPs, we employed varying depths of hidden layers.

## III. EXPERIMENTAL SETUP

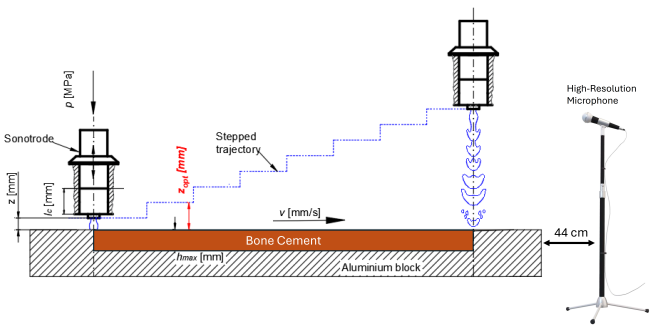

Fig. 2. Experimental setup showcasing the stairs trajectory used for varying standoff distance in increments of 20-70 mm. The bone cement is cut by a fluid jet inserted into an aluminum plate cavity, moving at a speed of 5 mm/s with each step taking 2 seconds, followed by a 1-second transition to the next step.

### A. Material Preparation

Palacos bone cement is prepared for the experiments by mixing its two components: the polymer powder and the monomer liquid, in accordance with the manufacturer's instructions, using hand mixing in a ceramic bowl under normal room conditions. For ease of handling and fixation, the cement dough was spread onto an aluminum matrix cavity and allowed to solidify. Once solidified, the sample was ready for testing.

### B. Experimental Configuration

The experiments are conducted at the Institute of Geonics, Ostrava, Czech Republic, utilizing ultrasonically excited pulsating fluid jet technology. This technology operates by forcing the decay of continuous high-pressure flow of the fluid via a high-frequency vibrating sonotrode (see Figure 3, A).

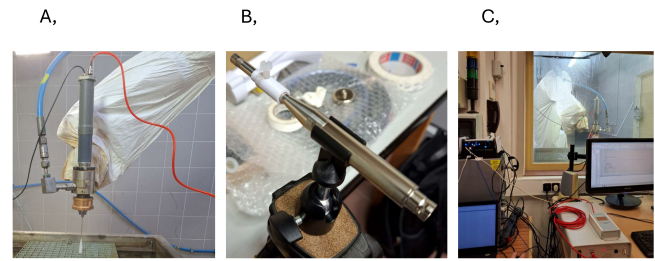

Fig. 3. Experimental setup for monitoring bone cement removal using a fluid jet with sonotrode (A), an audio monitoring microphone (B), and the overall setup view showing audio output channels on the left and screen display for the control of the robot arm on the right (C).

The study focuses on the effect of standoff distance $z$ (ranging from 2 to 7 mm) on the erosion profile while maintaining constant parameters: nozzle diameter $d = 0.6$ mm, sonotrode frequency $f = 22.17$ kHz (see Figure 4), supply pressure of the fluid $p = 20$ MPa, and robot arm velocity $v = 5$ mm/s. These parameters are selected based on optimal ranges identified in previous studies [7].

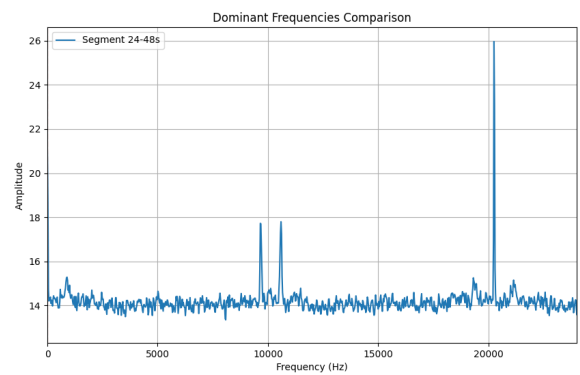

Fig. 4. Plot demonstrating the dominant frequency with pulsation, but without erosion of bone cement, that reflects the sonotrode frequency $f = 22.17$ kHz.

A stair trajectory is employed to vary the standoff distance, each disintegration segment having a length of 10 mm. Throughout, a highly sensitive acoustic microphone monitored the pulsating fluid jet to capture the audio data for further analysis (see Fig. 2).

The flexibility of the pulsating fluid jet and S4D-Bio model can be transferred beyond its current application. By modifying the system parameters, the methodology could be adapted for diverse material types and surgical settings. Specifically, surgical specializations that necessitate non-invasive precision, such as ophthalmology or cardiovascular procedures, might benefit from a similar approach.

### C. Data Collection and Analysis

The disintegration depth response is scanned with an optical microscope, converting the scanned data to digital form via MountainsLab software separated into x-, y- and z-axis values. This software extracted continuous depth profiles for each experimental condition. To improve statistical reliability, five equispaced continuous depth profiles parallel to the jet footprint were measured, resulting in a total of ten profiles per standoff distance for analysis and correlation. The grooves were created in two repetitions.

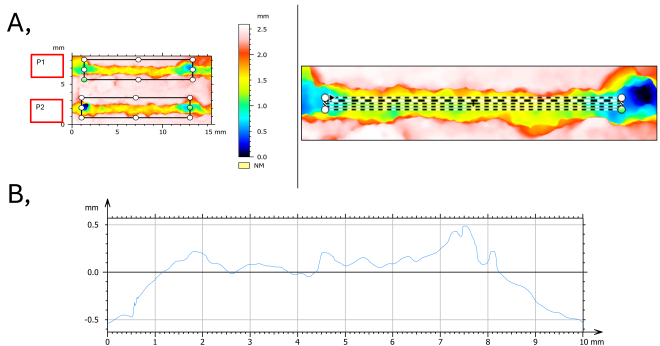

Fig. 5. (A) Erosion profiles showing material removal at different standoff distances for samples P1 and P2. (B) Cross-sectional displacement over a single trajectory.

The integration of a specialized acoustic recording system (see Figure 3, B) with our state space model was designed for real-time analysis, resulting in a cyber-physical system that combines audio data acquisition with predictive machine learning models.

## IV. Dataset Description

The dataset consists of numerical data representing ablation profiles as well as the audio files from the high resolution microphone. The ablation profiles are recorded in a CSV file with 726 MB size and input dimensions of [1150,70]. The measured erosion profiles of the pulsating fluid jet cut in the bone cement sample consists of five measurements for each erosion step to represent the variance of material erosion during fluid jet impact and of the FRT Scan itself. It secondly consists of two audio files, that have been recorded by a high resolution microphone with 38,4 kHz sampling rate and a distance of 44 cm between the microphone and the sample. They have a size of 26,4 MB and 86,7 MB and input dimensions of [1150, 60]. These audio files capture the sound of the fluid jet impacting the target and has been recorded in HDF5 format and being reformatted into wav. These audio files are preprocessed as Mel Spectrograms with 60 mel frequency bins for analyzing frequency distribution, since it has been an efficient way of preprocessing for audio data [11]. As the audio files start with running over the metal plate, the material properties of both materials, the metal as well as the bone cement are captured within the Mel Spectrograms (see Figure 6). These are cut to the equivalent length than the profile data and synchronized over time. In concatenated form we get input dimensions of [1150, 130].

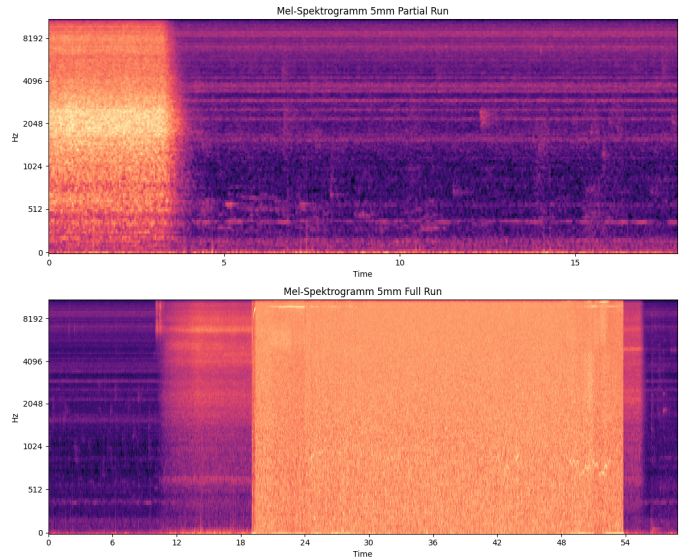

Fig. 6. The plot displays two Mel-spectrograms visualizing the removal process of bone cement (in orange) and the contact with metal (in purple). The vertical axis represents the frequency in Hertz, while the horizontal axis shows time in seconds. The color intensity indicates the amplitude: brighter tones correspond to higher amplitudes, while darker tones indicate lower amplitudes.

### A. Data Splitting

The data was divided into training and testing subsets. The initial 50% of the ablation profiles, accompanied by their corresponding audio data, were allocated to train the model. During the training phase, the known profiles and corresponding audio signal data, processed as Mel Spectrograms, serve as the model inputs. The model's parameters are optimized using the Mean Squared Error (MSE) loss function to minimize the prediction error between the predicted and actual erosion profiles. Once trained, the model predicts the remaining 50% of the ablation profiles, using audio data as input and comparing predictions against withheld ground-truth profiles.

Once the model has been trained on this initial dataset, it is then tasked with predicting the remaining 50% of the ablation profiles. The audio data corresponding to these profiles is used as the model's inputs, while the profiles themselves are withheld for validation purposes. Thus the erosion profiles are taken as ground-truth to evaluate the prediction accuracy.

## V. Predictive Modelling with S4D-Bio Model

In this study, we utilize continuous State Space Models (SSMs) to process acoustic signals for predictive modeling. SSMs are parameterized maps on signals represented either through linear ordinary differential equations (ODEs) or convolutions. These are expressed as follows:

$$x'(t) = Ax(t) + Bu(t) \\ y(t) = Cx(t) \quad (1)$$

$$K(t) = Ce^{tAB}, \quad y(t) = (K * u)(t) \quad (2)$$

State space models have been explored in dynamic system modeling due to their ability to accurately capture temporal dependencies in complex systems [39]. The adaptation of these models for surgical monitoring represents a novel application. The input dimensions of the dataset are [1150, 130]. The linear encoder layer transforms the full input from a dimension of 130 to the value determined by hyperparameter tuning with its best result at 256. According to hyperparameter tuning the number of S4D blocks varies from 1 to a possible maximum of 6. These blocks include the S4D cells, where each is given the dimension 256 for both input and output. This dimension represents the internal feature space of the model. The decoder transforms the hidden dimensions from 256 back to the output dimension of 70, corresponding to the output dimension of the erosion profiles. The primary task of the S4D-Bio model is to transform data through its hidden layers, efficiently processing the represented internal states while learning and enhancing the relevant signal patterns derived from the experimental data.

In diagonal SSMs (see Figure 7) like used in S4D-Bio, the state matrix $A$ is diagonal which allows for simplified computations by isolating the diagonal entries for efficient kernel calculation. In this context, the matrices $A$, $B$, and $C$ have dimensions according to the hyperparameter tuning, aligning with the input and output dimensionalities derived from the Mel Spectrogram shapes and the profile data. The model's parameters are structured to optimize the convolution operations [15], resulting in a single output value representing the predicted erosion profile.

The study also incorporates concepts from structured state spaces like the S4 model. Initially, the HiPPO-LegS framework provided a structured approach using the HiPPO-LegS matrix, though complex. Simplification comes through the S4 parameterization, where $A$ is decomposed into a normal matrix plus a rank-1 adjustment [15]:

$$A = A^{(N)} - PP^\top \quad (3)$$

The continuous SSMs are discretized through standard integration techniques like the bilinear transform and zero-order hold (ZOH) technique, transforming the continuous model into discrete approximations as shown:

$$y = u * K \quad \text{where} \quad K = (CB, CAB, \ldots, CA^{L-1}B) \quad (4)$$

For diagonal matrices, computing the convolution kernel is mathematically trivial, involving a Vandermonde matrix [15] (see Figure 7):

$$K_\ell = \sum_{n=0}^{N-1} C_n A_n^\ell B_n \Rightarrow K = (B^\top \circ C) \cdot VL(A) \quad (5)$$

This matrix $VL(A)$ is formulated to perform efficient computations in $O(N + L)$ operations and space.

## VI. RESULTS

The performance of the pulsating fluid jet device was evaluated through controlled experiments. These experiments aimed to determine the efficacy of the fluid jet in removing bone cement under varying standoff distances and audio feedback for process optimization.

### A. Efficacy of Long Nozzles

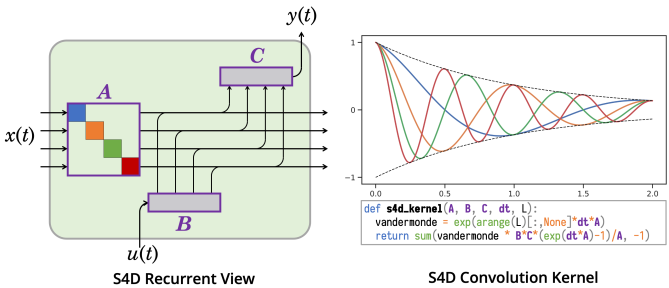

Fig. 7. S4D is a simplified diagonal state space model that retains the advantages of S4. On the left, the diagonal configuration enables its interpretation as a series of 1-dimensional state space models. On the right, as a convolutional framework, S4D features an easily understood convolution kernel (see Figure 7), which can be executed with minimal coding effort. The colors represent separate 1-D state space models, with purple indicating trainable parameters [15].

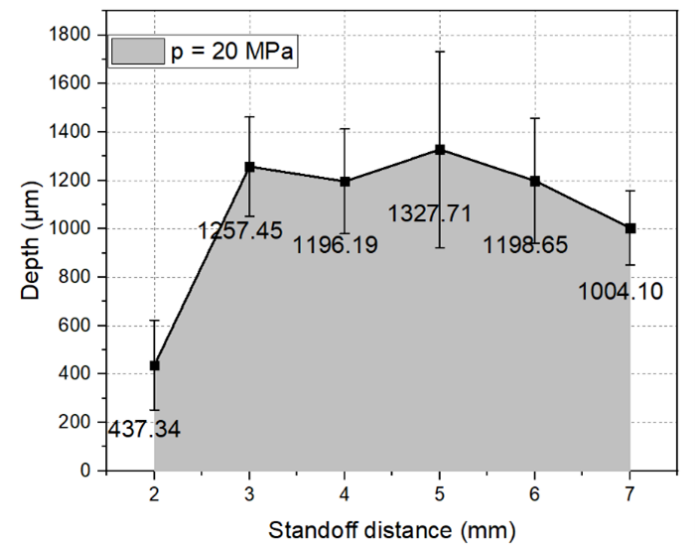

Fig. 8. Effect of standoff distance (z = 2 − 7 mm) on the disintegration depth at supply pressure p = 20 MPa.

The experiments confirmed the effectiveness of long nozzles in facilitating minimally invasive procedures. The erosion efficiency of the Fluidjet for the disintegration of bone cement was assessed in terms of the disintegration depth in the present study. The depth values used for the graphs are taken out of 5 equidistance cross-sections spread over the entire disintegrated length of 10 mm. 2 disintegration grooves were created at the specified supply pressure level (p = 20 MPa) and traverse speed (v = 5 mm/s) which makes a total of 10 disintegration depth values considered for the evaluation. It can be observed from Figure 8 that with an increase in the standoff distance, the disintegration depth increases. For instance, with an increase in the standoff distance, z = 2 to 3 mm, the disintegration depth increased from h = 437.34 ± 185.99 µm to 1257.45 ± 205.11 µm. This increase in the depth can be attributed to the morphology of the pulsating jet, as described in detail in a previous study [7]. When standoff distance increases from an initial level to higher distances, the morphology of the jet also changes. At a very short standoff distance, the propagation of the oscillating waves is not much developed, and the jet acts as continuous and the stagnation nature of the jet overpowers the impact nature of the water droplets. With the increase in the distance, the water clusters start forming, and the impact pressure regimes prevail over the stagnation pressure, which leads to an increase in the disintegration depth. This increase in the depth magnitude increases until a maximum limit corresponding to the optimal standoff distance. At the optimal standoff distance, discrete water droplet clusters are formed, inducing repetitive loading of the material and leading to fatigue failure. This phenomenon increases the efficiency of the jet, generating deeper erosion depth and is termed as culmination erosion stage. In the present study, maximal disintegration depth is observed to be h = 1327.71 ± 405.06 µm for standoff distance z = 5 mm. With further increase in the standoff distance, the aerodynamic effect between the jet and the surrounding takes place, deconcentrating the radial discrete water clusters leading to lesser erosion efficiency. This stage of erosion is termed the depletion stage and can be observed from z = 6 to 7 mm, with disintegration depth decreasing from 1198.65 ± 259.41 µm to 1004.1 ± 154.17 µm. Overall, it was observed that the erosion depth follows a bell-shaped curve with increasing standoff distance values. Furthermore, this trend of the depth is dependent on the ultrasonic and hydraulic factors, which were kept constant in the present study. Therefore, to achieve maximum efficiency in the process, the proper selection of standoff distance is crucial. Moreover, it is also evident that the disintegration process is not very uniform and depends on the initial surface or the purity of the samples. These irregularities in the depth results are evident from the larger error bars associated with the mean values for every standoff distance. As pointed out, it can be due to the material preparation, which in this study can be during the mixing of the components of the bone cement in normal laboratory conditions, which could have generated voids inside the material, which acts as an initiator site for stress concentration when impacted with the Fluidjet. Subsequently, the cracks start from these weak points and propagate throughout the material leading to bulk removal of the material sections.

### B. Machine Learning Results

Our data analysis reveals distinct audio patterns that are strongly correlated with cement removal. The S4D-Bio model yielded an validation accuracy of 98.93 % in predicting the erosion profiles from synchronously recorded audio data. Here, accuracy is defined as the percentage of correctly predicted points within an acceptable error threshold of 1 $\mu$m. This definition is chosen to emphasize pattern recognition capabilities rather than micrometer-level precision. This demonstrates the efficacy of state space models in capturing temporal dynamics and material properties as the S4D-Bio model is able to learn the relationship between the audio signatures and the erosion profile depths with minimal error, indicating its potential in improving monitoring systems within orthopedic surgeries.

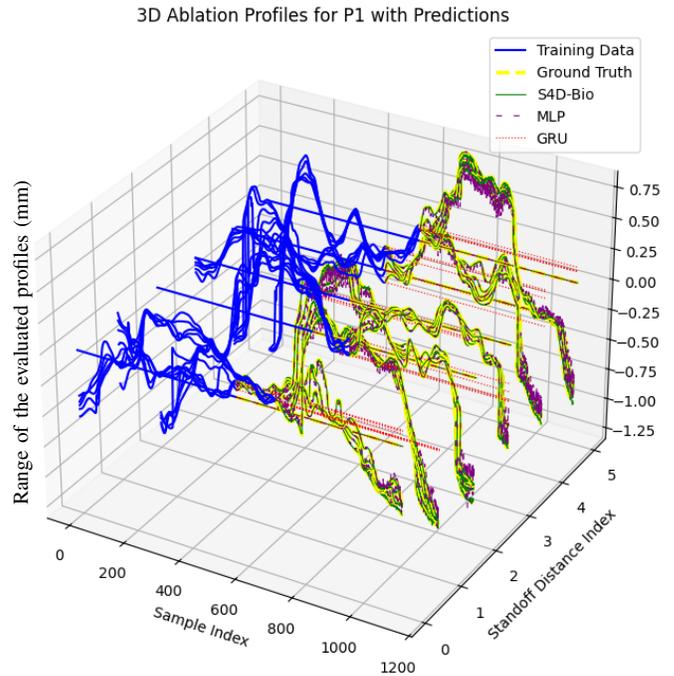

Fig. 9. Predicted profiles from the benchmarking models (see legend) versus trained data (blue) and ground truth (yellow).

Figure 9 presents the 3D visualization of ablation profiles, comparing the trained model's profile (blue) with ground truth (yellow) and the model's predicted profiles (green for S4D-Bio, dashed purple for MLP and red for GRU). The x-axis represents the sample points, while the y-axis shows the Standoff Distance arranged according to the experimental setup's stepped trajectory, and the z-axis depicts the disintegration distance of bone cement. The high degree of overlap between the S4D-Bio model's green predicted profile and the yellow ground truth curve shows the model's efficacy in replicating the intricate erosion patterns typically observed during orthopedic interventions, while the GRU model seems to achieve low accuracy as it repeats to predict the last value it gets from training data.

### C. Results in the Handling of Noise

The presence of noise in data collection is an important factor that can significantly impact the performance of machine learning models, particularly in engineering and real-world applications [40, 41]. In deep learning, substantial progress has been made in addressing noise issues through various end-to-end solutions. Techniques such as robust training, noise filtering or adversarial training have been developed to enhance model performance when faced with noisy input data [42, 43].

In the context of this fluidjet application with its audio monitoring, noise originates from various sources, including environmental factors, equipment vibrations, the intentionally uneven bone cement surfaces and background interference. This noise can potentially mask important audio signals that are used to monitor and control the surgical process, leading to inaccuracies in predictions and model outputs. Therefore, techniques such as dropout [44] and batch normalization [45] are employed during the training process to improve the model's resilience against noisy inputs. As can be observed from the high accuracy values in Table I S4D-Bio can be regarded as a noise-tolerant architecture, that can be used to learn these effects in an end-to-end manner while applying these techniques.

While these approaches to automatically denoise the data in an end-to-end manner are becoming state-of-the-art in the field of machine learning, their application in engineering disciplines, particularly in biomedical engineering, is still evolving. By leveraging these noise-handling strategies, the efficacy and reliability of machine learning models in environments with high noise levels, such as surgical settings, can be significantly improved. This integration represents not only a technical advancement but also contributes to the broader field of engineering by promoting cross-disciplinary innovations.

### D. Hyperparameter Optimization results

We use optuna [46] to optimize the hyperparameters with AutoML. The number of trials is set to 50 and the number of trained epochs is set to 30 for each model in order to have a fair comparison. Dropout [47] is optimized for all models equally and Adam [48] is used as the optimizer for all models. The model-dimensions, the number of blocks as well as the learning-rate are also optimized for all models as well as the number of hidden layers and the hidden size for the Deep-MLP. All models are trained on the same CPU with 4 workers.

### E. Benchmarking results

The experiment demonstrated that State Space Models are more adaptable to varying step sizes in the data compared to Long Short-Term Memory [36] and gated recurrent units [37]. SSMs provide a more robust alternative, handling data with varying temporal resolutions. As can be shown from Figure 9 the GRU tends to predict the last values from the training data and therefore gains a bad prediction accuracy. The same can be regarded for LSTM's. As both are originally developed for the prediction of time-series and the profiles of this dataset are rather spatio-temporal data, this could be the reason for their bad correlation. Small thresholds are considered acceptable error ranges at each point along the erosion profile. Thus, in this context, accuracy is defined here as the percentage of correctly predicted points within an error threshold of $1\,\mu$m. This choice of metric allows us to focus on the model's ability to correctly identify the pattern of erosion rather than absolute precision, making it especially suitable for applications involving spatio-temporal profile prediction. Since the goal in this application is to identify the correct pattern rather than achieving micrometer-level precision at each point, we use accuracy as our primary metric for model evaluation [49]. This approach allows us to benchmark models based on their pattern recognition capability, making accuracy more suitable for this task than MSE or RMSE.

TABLE I
BENCHMARKING MODELS AND THEIR ACCURACY

| Model | Accuracy (%) |
|---|---|
| **S4D-Bio** | **98.93** |
| GRU | 0.03 |
| (Shallow)-MLP | 90.75 |
| Deep-MLP | 94.99 |

### F. Limitations and Considerations

To ensure the safety and effectiveness of the pulsating fluid jet technology, these initial experiments are conducted under controlled laboratory conditions. This approach is essential to prevent any risk to patients during early development stages and contributes significantly to foundational research in surgical innovation. While the results are promising, areas for further research include optimizing nozzle configuration for different cement types and bone structures under surgical scenarios. The acoustic monitoring technique requires further refinement to accommodate varying surgical environments. Additionally, investigations into potential tissue interactions at varying frequencies are warranted to ensure safe application across diverse procedures.

## VII. CONCLUSIONS

We demonstrate the feasibility and effectiveness of utilizing pulsating fluid jet technology, integrated with advanced audio monitoring techniques, for the minimally invasive and cold removal of bone cement in orthopedic applications under laboratory conditions. The applied state space model S4D-Bio, shows achieves a validation accuracy of 98.93% in correlating audio data with erosion profiles. This represents a significant advancement in surgical precision and the potential for real-time adjustment during procedures.

Our experiments confirm the efficacy of long nozzles, which are instrumental in facilitating minimally invasive surgical approaches. Specifically, the use of long nozzles allows for precise targeting of bone cement removal. The erosion depth follows a bell-shaped curve concerning the standoff distance, highlighting the critical nature of optimizing procedural parameters to maximize efficiency.

Audio monitoring, an unexplored technique in the domain of pulsating fluidjets, proves to be feasible in overcoming the traditional challenge of visibility obstruction due to splashing effects. By capturing distinct acoustic signatures during the cutting process, the S4D-Bio model adeptly differentiates between different bone cement erosion rates.

48     IEEE TRANSACTIONS ON BIOMEDICAL ENGINEERING, VOL. XX, NO. XX, XXXXX 2023The methodological framework developed in this research has broader implications beyond its immediate application. The integration of noise-handling strategies, particularly dropout and batch normalization during the training process, demonstrates to enhance the models robustness in noisy environments and can be used as starting-point for further cross-disciplinary research in biomedical engineering.

The successful adaptation of the S4D-Bio framework for surgical monitoring paves the way for future enhancements and adaptations across various medical and industrial fields. Additionally, this study lays the groundwork for subsequent exploration of the fluid hammer effect as a subprocess—an endeavor that could yield even greater insights into the spatial-temporal complexities of material erosion.

Future work will also focus on in vivo validation of these results to cater to diverse surgical needs ensuring safer and more effective patient outcomes.

## VIII. Acknowledgments

This work is partly financed by the Federal Ministry of Education and Research over the OrthoJet Project with grant number 13GW0586F. We thank our project partners for their valuable insights.## References

[1] İbrahim Kapici and Fende Sermin Utku. "A study on injectable bone cement". In: *2017 Medical Technologies National Congress (TIPTEKNO)*. 2017, pp. 1–4. DOI: 10.1109/TIPTEKNO.2017.8238059.

[2] Gabriele Panegrossi et al. "Bone loss management in total knee revision surgery". In: *International orthopaedics* 38 (2014), pp. 419–427.

[3] Wayne G Paprosky, Steven H Weeden, and Jack W Bowling Jr. "Component removal in revision total hip arthroplasty". In: *Clinical Orthopaedics and Related Research®* 393 (2001), pp. 181–193.

[4] M Zimmer et al. "Bone-cement removal with the excimer laser in revision arthroplasty". In: *Archives of orthopaedic and trauma surgery* 112 (1992), pp. 15–17.

[5] Julia Schnieders et al. "Controlled release of gentamicin from calcium phosphate—poly (lactic acid-co-glycolic acid) composite bone cement". In: *Biomaterials* 27.23 (2006), pp. 4239–4249.

[6] M. Aguas et al. "A Novel Technique to Remove Bone Cement in Reoperative Revision Knee Arthroplasty". In: *2013 39th Annual Northeast Bioengineering Conference*. 2013, pp. 241–242. DOI: 10.1109/NEBEC.2013.43.

[7] Sergej Hloch et al. "Disintegration of bone cement by continuous and pulsating water jet". In: *Tech. Gaz* 20.4 (2013), pp. 593–598.

[8] Akash Nag et al. "Utilization of ultrasonically forced pulsating water jet decaying for bone cement removal". In: *The International Journal of Advanced Manufacturing Technology* 110 (2020), pp. 829–840.

[9] Matthias Honl et al. "The use of water-jetting technology in prostheses revision surgery—First results of parameter studies on bone and bone cement". In: *Journal of Biomedical Materials Research: An Official Journal of The Society for Biomaterials, The Japanese Society for Biomaterials, and The Australian Society for Biomaterials and the Korean Society for Biomaterials* 53.6 (2000), pp. 781–790.

[10] M.C. Zimmerman et al. "The evaluation of cortical bone remodeling with a new ultrasonic technique". In: *IEEE Transactions on Biomedical Engineering* 37.5 (1990), pp. 433–441. DOI: 10.1109/10.55634.

[11] Pascal Janetzky et al. "Swarming Detection in Smart Beehives Using Auto Encoders for Audio Data". In: *2023 30th International Conference on Systems, Signals and Image Processing (IWSSIP)*. 2023, pp. 1–5. DOI: 10.1109/IWSSIP58668.2023.10180253.

[12] Sergej Hloch et al. "On-line measurement and monitoring of pulsating saline and water jet disintegration of bone cement with frequency 20 kHz". In: *Measurement* 147 (2019), p. 106828.

[13] Pavol Hreha et al. "Water jet technology used in medicine". In: *Tehnicki vjesnik* 17.2 (2010), pp. 237–240.

[14] M. Sarlinova et al. "Fracture properties of bone cements". In: *2014 ELEKTRO*. 2014, pp. 616–620. DOI: 10.1109/ELEKTRO.2014.6848972.

[15] Ankit Gupta, Albert Gu, and Jonathan Berant. *Diagonal State Spaces are as Effective as Structured State Spaces*. 2022. arXiv: 2203.14343 [cs.LG].

[16] A. M. C. Lopes et al. "Characterization of Thermal Properties of Polymethylmethacrylate Bone Cement for Therapeutic Ultrasound Application". In: *2023 Global Medical Engineering Physics Exchanges/Pacific Health Care Engineering (GMEPE/PAHCE)*. 2023, pp. 1–3. DOI: 10.1109/GMEPE/PAHCE58559.2023.10226396.

[17] Pavol Hvizdoš et al. "Local mechanical properties of various bone cements". In: *Key Engineering Materials* 592 (2014), pp. 382–385.

[18] WW Duncan et al. "Revision of the cemented femoral stem using a cement-in-cement technique: a five-to 15-year review". In: *The Journal of Bone & Joint Surgery British Volume* 91.5 (2009), pp. 577–582.

[19] BH Moeckel et al. "Cement-within-cement revision hip arthroplasty". In: *The Journal of Bone & Joint Surgery British Volume* 75.6 (1993), pp. 869–871.

[20] A Liddle et al. "Ultrasonic cement removal in cement-in-cement revision total hip arthroplasty: what is the effect on the final cement-in-cement bond?" In: *Bone & joint research* 8.6 (2019), pp. 246–252.

[21] Matthias Honl et al. "The water jet as a new tool for endoprothesis revision surgery–an in vitro study on human bone and bone cement". In: *Bio-medical materials and engineering* 13.4 (2003), pp. 317–325.

[22] Peter Helwig et al. "Tibial cleaning method for cemented total knee arthroplasty: An experimental study". In: *Indian journal of orthopaedics* 47 (2013), pp. 18–22.